\documentclass{ifacconf}

\usepackage{graphicx}      
\usepackage{natbib}  
\usepackage{amsmath}
\usepackage{booktabs}
\begin{document}
\begin{frontmatter}

\title{Improved YOLOv12 with LLM-Generated Synthetic Data for Enhanced Apple Detection and Benchmarking Against YOLOv11 and YOLOv10\thanksref{footnoteinfo}} 

\thanks[footnoteinfo]{Sponsor and financial support acknowledgment
goes here. Paper titles should be written in uppercase and lowercase
letters, not all uppercase.}

\author[First]{Ranjan Sapkota}
\author[Second,Third]{Manoj Karkee}

\address[First]{Cornell University,
Department of Environmental and Biological Systems Engineering,
Ithaca, NY 14853, USA }
\address[Second]{Cornell University,
Department of Biological and Environmental Engineering,
Ithaca, NY 14853, USA}
\address[Third]{Washington State University,
Department of Biological Systems Engineering,
Center for Precision \& Automated Agricultural Systems,
Prosser, WA 99350, USA (e-mail: manoj.karkee@wsu.edu)}

\begin{abstract}                
This study evaluated the performance of the YOLOv12 object detection model, and compared against the performances YOLOv11 and YOLOv10 for apple detection in commercial orchards based on the model training completed entirely on synthetic images generated by Large Language Models (LLMs). The YOLOv12n configuration achieved the highest precision at 0.916, the highest recall at 0.969, and the highest mean Average Precision (mAP@50) at 0.978. In comparison, the YOLOv11 series was led by YOLO11x, which achieved the highest precision at 0.857, recall at 0.85, and mAP@50 at 0.91. For the YOLOv10 series, YOLOv10b and YOLOv10l both achieved the highest precision at 0.85, with YOLOv10n achieving the highest recall at 0.8 and mAP@50 at 0.89. These findings demonstrated that YOLOv12, when trained on realistic LLM-generated datasets surpassed its predecessors in key performance metrics. The technique also offered a cost-effective solution by reducing the need for extensive manual data collection in the agricultural field. In addition, this study compared the computational efficiency of all versions of YOLOv12, v11 and v10, where YOLOv11n reported the lowest inference time at 4.7 ms, compared to YOLOv12n's 5.6 ms and YOLOv10n's 5.9 ms. Although YOLOv12 is new and more accurate than YOLOv11, and YOLOv10,  YOLO11n still stays the fastest YOLO model among YOLOv10, YOLOv11 and YOLOv12 series of models. 
\end{abstract}


\end{frontmatter}

\section{Introduction}
Robust object detection is crucial for advancing agricultural automation, particularly in complex environments like apple orchards. Traditionally, apple detection has relied on field imaging using expensive sensors and labor-intensive setups to capture diverse conditions, including occlusions, variable lighting, and dense foliage \cite{dhanya2022deep,tian2020computer}. This approach not only increases costs but also limits dataset diversity, hindering the development of deep-learning models. Recent advancements in apple detection have introduced various methodological innovations. For instance, \cite{liu2024faster} developed Faster-YOLO-AP, a lightweight model using efficient PDWConv for high speed and accuracy on edge devices. \cite{johanson2024s3ad} proposed a semi-supervised approach (S3AD) leveraging mixed labeled and unlabeled datasets to improve small apple detection. \cite{ma2024using} integrated ShuffleNetv2 and Ghost modules into a lightweight YOLOv8 variant for real-time monitoring, achieving high efficiency and precision. \cite{kong2024detection} enhanced detection with a transformer-based Faster R-CNN model, excelling in complex orchard environments. \cite{jin2025enhanced} optimized  YOLOv8n for robotic harvesting, achieving good localization and counting accuracy. Similarly, \cite{maheswari2025performance} analyzed a modified DeepLabv3+ architecture for high-accuracy fruit localization. While these studies represent significant progress, they often incur high costs due to sensor-based image collection, labor-intensive processes, and manual annotation.

While previous studies have refined YOLO-based detectors, such as the lightweight Faster-YOLO-AP for occlusion-heavy orchards \cite{liu2024faster} and enhanced YOLOv8 for multi-stage fruit detection under variable conditions \cite{ma2024using}—they predominantly rely on field-collected data. Other works, including attention-enhanced Faster R-CNN \cite{kong2024detection} and semi-supervised frameworks like S3AD \cite{johanson2024s3ad}, have sought to mitigate false positives and improve small-object detection. However, these approaches still face challenges related to dataset diversity and high computational costs for edge deployment. 

Recent advances in generative AI, particularly with Large Language Models (LLMs) such as OpenAI’s DALL-E, have transformed synthetic image generation. By converting simple textual prompts (e.g., ``occluded apples in orchards'') into photorealistic images, LLMs enable the creation of scalable, annotated datasets that faithfully replicate real-world complexities without the need for physical data collection \cite{sapkota4941582synthetic}. This advancement addresses critical limitations imposed by traditional sensor-based imaging systems. 

Likewise, the YOLO family of detectors has long been favored for its superior speed-accuracy trade-off in object detection tasks \cite{redmon2016you,wang2024yolov10,wang2024yolov9}. Although recent iterations such as YOLOv10, YOLOv11, and YOLOv12 have achieved notable improvements, their reliance on manually curated datasets continues to limit their scalability and robustness under diverse environmental conditions. By integrating LLM-generated synthetic data, training scenarios become infinitely customizable, enhancing model robustness to real-world variability.

Building on our previous work \cite{sapkota4941582synthetic}, where we demonstrated the feasibility of generating synthetic datasets using OpenAI's DALL-E for training YOLOv10 and YOLOv11 models, this study introduces \textbf{YOLOv12} (Architecture in Figure \ref{fig:IFAC1}), a state-of-the-art model that leverages the scalability of LLM-generated datasets and advanced architectural innovations. In our earlier research, we showed that LLM-generated datasets have the potential to replace traditional field data collection, achieving high precision, recall, and mAP@50 with YOLOv11 and YOLOv10 models. Its nano variant (YOLOv12-N) achieves an inference speed of 1.64 ms on a T4 GPU, making it highly suitable for real-time applications like robotic harvesting  \cite{tian2025yolov12}. By eliminating costly sensor setups and extensive field labor, our approach overcomes traditional limitations and demonstrates superior generalization across occlusion, lighting, and scale variations \cite{jha2019comprehensive,meng2025yolov10}. The specific objectives of this study are as follows:

\begin{itemize}
    \item Prepare LLM-generate synthetic datasets for training four configurations (n, s, m and l) of YOLOv12 model and assess  their performances on synthetic data. 
    \item Compare the performance of YOLOv12 models with different configurations of YOLOv11 and YOLOv10 models using the same synthetic datasets.
    \item     \textbf{Field-Level Testing with Real Images:} To validate the trained models using real-world images collected by machine vision sensors in commercial apple orchards to test the practical applicability of the models in real agricultural settings.
\end{itemize}

\section{Methods}
\begin{figure*}[ht]
\centering
\includegraphics[width=0.99\linewidth]{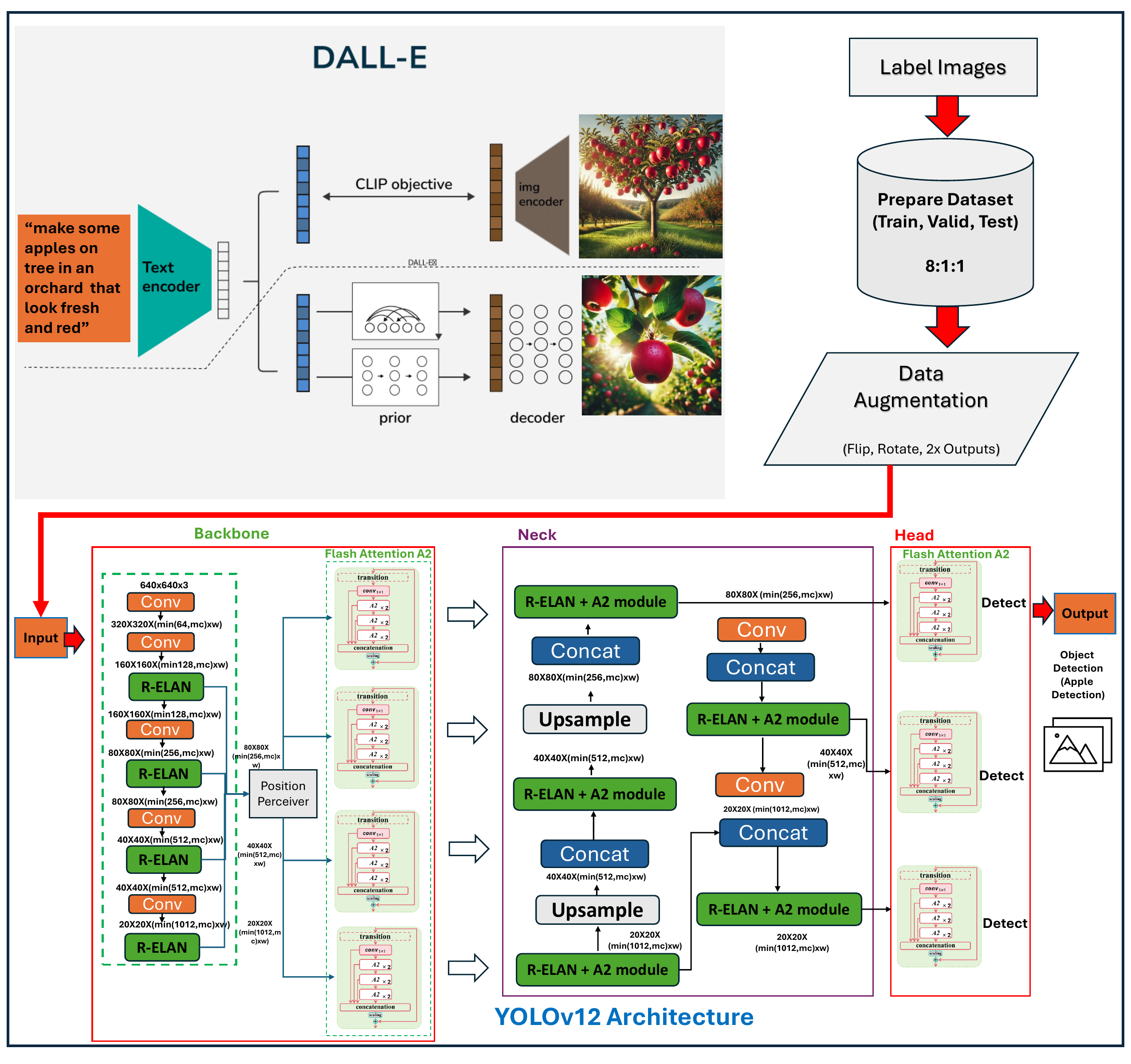}
\caption{ Summarizing the process of image generation to train YOLOv12 object detection model using synthetic images. Prompt engineering used to realistic image generation for this study is explained in our previous study \cite{sapkota2024yolov10}. Lower panel illustrates the architecture of YOLOv12 object detection model.}
\label{fig:IFAC1}
\end{figure*}

This study expanded on prior work by \cite{sapkota4941582synthetic}, which demonstrated the feasibility of training YOLOv10 and YOLOv11 models using synthetic datasets generated via OpenAI’s DALL-E as explained in Figure \ref{fig:IFAC1}.  For this research, the synthetic dataset consisting of 489 manually annotated images, each sized at 1024x1024 pixels, was prepared for training the YOLOv12 object detection model. The methodology can be summarized in four steps; i) Image Generation using Large Language Model and Dataset Preparation; ii)   Training YOLOv12 object detection model for detecting apples in synthetic images; iii) Performance Metrics Evaluation and Comparison with YOLO11 and YOLOv10,  and iv) Testing with real-field images collected by a machine vision sensor.

\subsection{LLM-based Data Generation and Preparation}
A Large Language Model (LLM) developed by OpenAI, The DALL·E 2, was used to generate synthetic orchard images through a hierarchical text-conditional strategy. This process involves two primary phases: (1) generating a CLIP (Contrastive Language-Image Pretraining) image embedding from text prompts; and (2) decoding this embedding into photorealistic, contextually relevant images \cite{ramesh2022hierarchical}. The pipeline comprises three stages: encoder, prior, and decoder. First, textual inputs (e.g., “make some apples on tree in an orchard that look fresh and red”) are converted into CLIP text embeddings using a pretrained neural network \cite{sapkota4941582synthetic}. These embeddings undergo dimensionality reduction via Principal Component Analysis (PCA) to streamline processing. In the prior stage, a Transformer model with attention mechanisms transforms the text embeddings into image embeddings \cite{paass2023foundation}. Finally, a diffusion-based decoder progressively refines the embeddings into high-resolution images, upscaled from 64×64 to 1024×1024 pixels through convolutional networks \cite{xing2024survey, paass2023foundation}. This method enables flexible image generation, adjusting to dynamic textual inputs while preserving semantic integrity, even for complex tasks like inpainting.  

Initially, 501 images were generated using diverse prompts. After quality filtering (\cite{sapkota4941582synthetic}), 20 unrealistic outputs (e.g., abstract or apple-free scenes) were discarded, resulting in 489 images depicting ripened apples in realistic orchard environments. These images were manually annotated using Makesense.ai, an open-source platform, to add bounding boxes for 8,590 apples. Post-annotation, images were standardized to 640×640 resolution and augmented using horizontal and vertical flips, and 90-degree rotations (clockwise, counterclockwise, and upside-down) to enhance dataset robustness.  Although DALL·E 2 generates synthetic images, it lacks native annotation capabilities, necessitating manual labeling. Additionally, the dataset was uploaded to Roboflow (Roboflow, Ames Iows, USA) and exported in YOLOv12 data format to further train YOLOv12 object detection model and its four configurations. 

\subsection{Training YOLOv12 Object Detection Model}
The YOLOv12 model was trained using a structured methodology to ensure reproducibility and performance optimization. The training process used \textbf{200 epochs} with an input image size of \textbf{640 pixels} and a batch size of \textbf{8}, balancing computational efficiency and feature learning. Training was conducted on a high-performance workstation equipped with an \textbf{Intel Xeon® W-2155 CPU} (3.30 GHz, 20 cores), an \textbf{NVIDIA TITAN Xp GPU}, \textbf{31.1 GiB RAM}, and an \textbf{Ubuntu 16.04 LTS 64-bit OS}. The PyTorch framework was employed for implementation, leveraging its compatibility with deep learning workloads on Linux systems.

Model checkpoints were saved every \textbf{10 epochs} in a dedicated directory, enabling retrospective evaluation of performance improvements. The training setup remained consistent across all YOLOv12 configurations (\textbf{n, s, m, l }), with uniform hyperparameters (e.g., optimizer settings, learning rate) to facilitate fair comparisons with predecessors (YOLOv11 and YOLOv10 in \cite{sapkota4941582synthetic}). Key architectural innovations, including the \textbf{Area Attention (A\textsuperscript{2}) Module} (50\% computational complexity reduction via spatial reshaping), \textbf{Residual ELAN} (18\% fewer parameters and 24\% fewer FLOPs), and \textbf{7×7 depth-wise convolution} (replacing positional encoding), were integrated into the training framework. These components collectively enhanced real-time detection capabilities and stability.

The model utilized \textbf{SGD with cosine learning rate scheduling} (initial \( \text{lr} = 0.01 \)) and incorporated augmentations like \textbf{Mosaic-9} and \textbf{Mixup}, which contributed to a \textbf{12.8\% mAP gain} on COCO benchmarks. Computational efficiency was further optimized through adaptive MLP ratios (1.2×) and shallow block stacking, achieving \textbf{4.1 ms inference latency} on NVIDIA V100 GPUs. All configurations maintained fixed-resolution processing (\( n = 640 \)) with optimized memory access patterns, ensuring scalability across hardware environments. This rigorous methodology enabled robust performance comparisons while minimizing variability in training conditions.

\subsection{Performance Evaluation}

To systematically assess the efficacy of the YOLOv12 model across its four configurations (n, s, m, and l), a consistent set of performance metrics was employed. These metrics included box precision, box recall, and mean average precision (mAP) at an Intersection over Union (IoU) threshold of 50\%. These evaluations were crucial for determining the model's accuracy and efficiency in detecting apples in LLM-generated, synthetic images. The formulas for these metrics are as follows:

\begin{equation}
Precision = \frac{TP}{TP + FP}
\end{equation}

\begin{equation}
Recall = \frac{TP}{TP + FN}
\end{equation}

In addition to accuracy metrics, the model's complexity and computational demand were evaluated by assessing the number of convolutional layers, total parameters, and GFLOPs:

\begin{equation}
Parameters_{Model} = \text{Total trainable weights and biases}
\end{equation}

\begin{equation}
GFLOPs = \frac{\text{Total floating-point operations}}{10^9} \text{ per image}
\end{equation}

\begin{equation}
Layers_{Convolutional} = \text{Total convolutional layers }
\end{equation}

\begin{figure}[ht]
\centering
\includegraphics[width=0.90\linewidth]{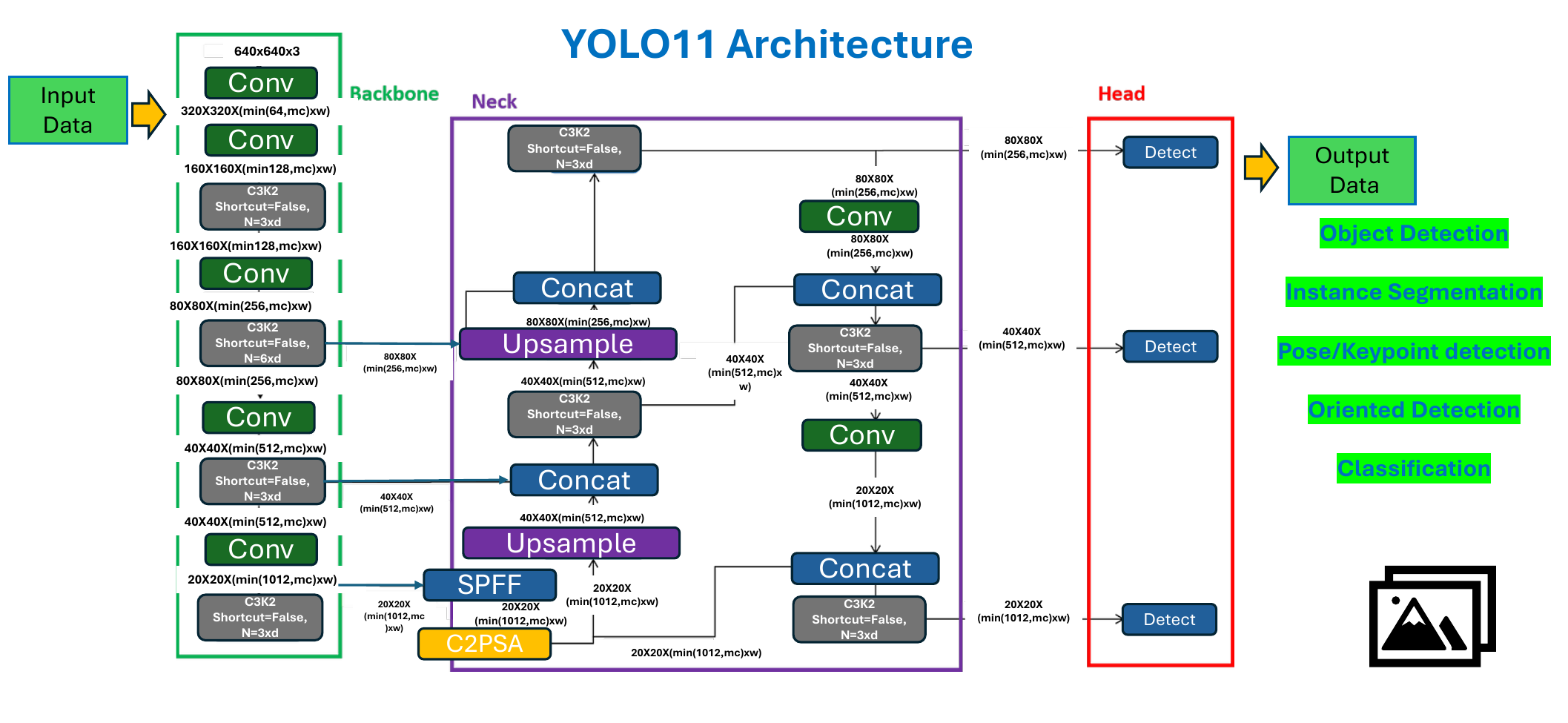}
\caption{ YOLOv11 Architecture Diagram}
\label{fig:ArchitectureYOLO11}
\end{figure}

\begin{figure}[ht]
\centering
\includegraphics[width=0.90\linewidth]{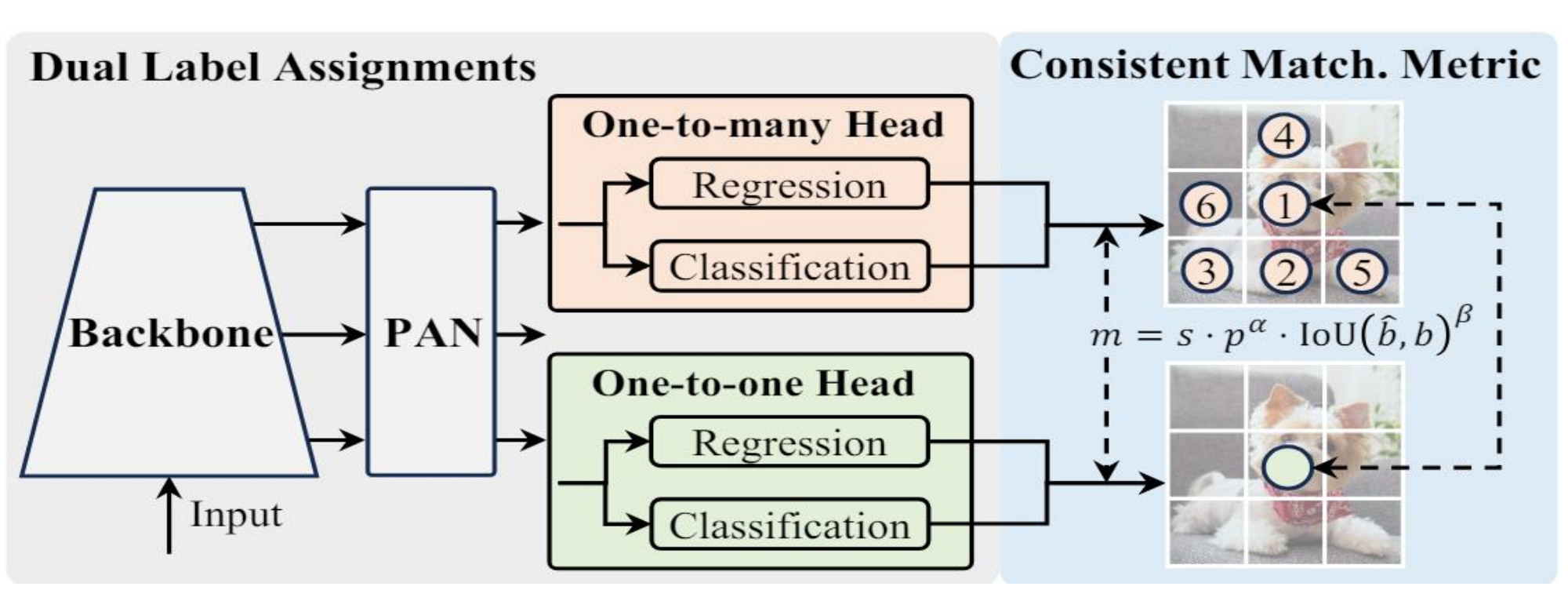}
\caption{ YOLOv10 Architecture Diagram}
\label{fig:ArchitectureYOLO10}
\end{figure}

These operational metrics  and image processing speeds (Inference) provide valuable information into the scalability and deployment feasibility of YOLOv12, particularly in applications requiring high-throughput and real-time processing. The performance of YOLOv12 was then compared against that of its predecessors, YOLOv11 (Figure \ref{fig:ArchitectureYOLO11} and YOLOv10 \ref{fig:ArchitectureYOLO10}, using the same dataset (and earlier experiment \cite{sapkota4941582synthetic}. This comparative analysis helps highlight the advancements in YOLOv12's design and its implications for practical applications in agricultural settings, especially for tasks involving the detection of apples in complex orchard environments.

\subsection{Field Testing with Real World Images}  
The YOLOv12 model, trained solely on LLM-generated synthetic data, was then deployed for inference testing (not formal validation) using real-world images captured in a Washington State apple orchard on September 29, 2024. To collect this dataset, a Microsoft Azure Kinect DK sensor—equipped with a 12MP RGB camera and a 1MP ToF depth sensor—was mounted on a UR5e robotic arm (Universal Robots) affixed to a Clearpath Warthog unmanned ground vehicle. Using this setup, 40 high-resolution (1920x1080) images of Scilate apples under field conditions. The sensor’s global shutter and configurable depth modes (NFOV/WFOV) ensured low-noise, and synchronized pixel capture. This field testing was essential to assess YOLOv12’s practical applicability in agricultural environments.  

\section{Results and Discussion}
\subsection{Apple Detection Performance}
In the comparative analysis of object detection models, YOLOv12 configurations demonstrated good performance in detecting apples on synthetic LLM-generated images. Among the YOLOv12 variants, the YOLOv12n configuration emerged as the most accurate, recording the highest box precision at 0.916, the highest box recall at 0.969, and the highest mAP@50 at 0.978. The YOLOv12s, YOLOv12m, and YOLOv12l models exhibited closely matched performance, each achieving a precision of 0.898, a recall of 0.956, and a mAP@50 of 0.974, demonstrating the consistency across these configurations.

Similarly, the YOLO11 and YOLOv10 family of model configurations also showed reasonably good  performance under similar conditions. Within the YOLO11 series, the YOLO11x configuration reached the highest precision of 0.857 and the highest mAP@50 of 0.91, while YOLO11m configuration achieved the highest recall at 0.821. Among the YOLOv10 configurations, YOLOv10n and YOLOv10b both achieved the highest precision of 0.85, with YOLOv10n achieving the highest mAP@50 at 0.89. YOLOv10x achieved the highest recall at 0.81. These results clearly show the progression and refinement in YOLO model development over time, with newer iterations showing relatively better accuracy. For a detailed breakdown of each model’s metrics and a comprehensive comparison between YOLOv12, YOLO11, and YOLOv10, please refer to Table \ref{tab:performance_metrics}, which presents the complete performance data for these configurations. The results, in summary, show that YOLOv12n is the superior model configuration among the tested ones in terms of precision, recall, and mAP.
\begin{table*}[ht]
\centering
\caption{\textbf{Comparative performance metrics for YOLOv12, YOLO11, and YOLOv10 models on synthetic LLM-generated apple detection images.}}
\label{tab:performance_metrics}
\begin{tabular}{@{}c|ccc@{}}
\toprule
\textbf{Model Configuration} & \textbf{Precision} & \textbf{Recall} & \textbf{mAP@50} \\
\midrule
\textbf{YOLOv12n} & \textbf{0.916} & \textbf{0.969} & \textbf{0.978} \\
\textbf{YOLOv12s} & 0.898 & 0.956 & 0.974 \\
\textbf{YOLOv12m} & 0.898 & 0.956 & 0.974 \\
\textbf{YOLOv12l} & 0.898 & 0.956 & 0.974 \\
\midrule
\textbf{YOLO11n}  & 0.84  & 0.76  & 0.862 \\
\textbf{YOLO11s}  & 0.874 & 0.826 & 0.909 \\
\textbf{YOLO11m}  & 0.809 & 0.821 & 0.879 \\
\textbf{YOLO11l}  & 0.836 & 0.877 & 0.866 \\
\textbf{YOLO11x}  & 0.857 & 0.85  & 0.91  \\
\midrule
\textbf{YOLOv10n} & 0.84  & 0.8   & 0.89  \\
\textbf{YOLOv10s} & 0.82  & 0.83  & 0.88  \\
\textbf{YOLOv10m} & 0.83  & 0.8   & 0.87  \\
\textbf{YOLOv10b} & 0.85  & 0.82  & 0.88  \\
\textbf{YOLOv10l} & 0.85  & 0.75  & 0.83  \\
\textbf{YOLOv10x} & 0.77  & 0.81  & 0.85  \\
\bottomrule
\end{tabular}
\end{table*}

\begin{figure*}[ht]
\centering
\includegraphics[width=0.85\linewidth]{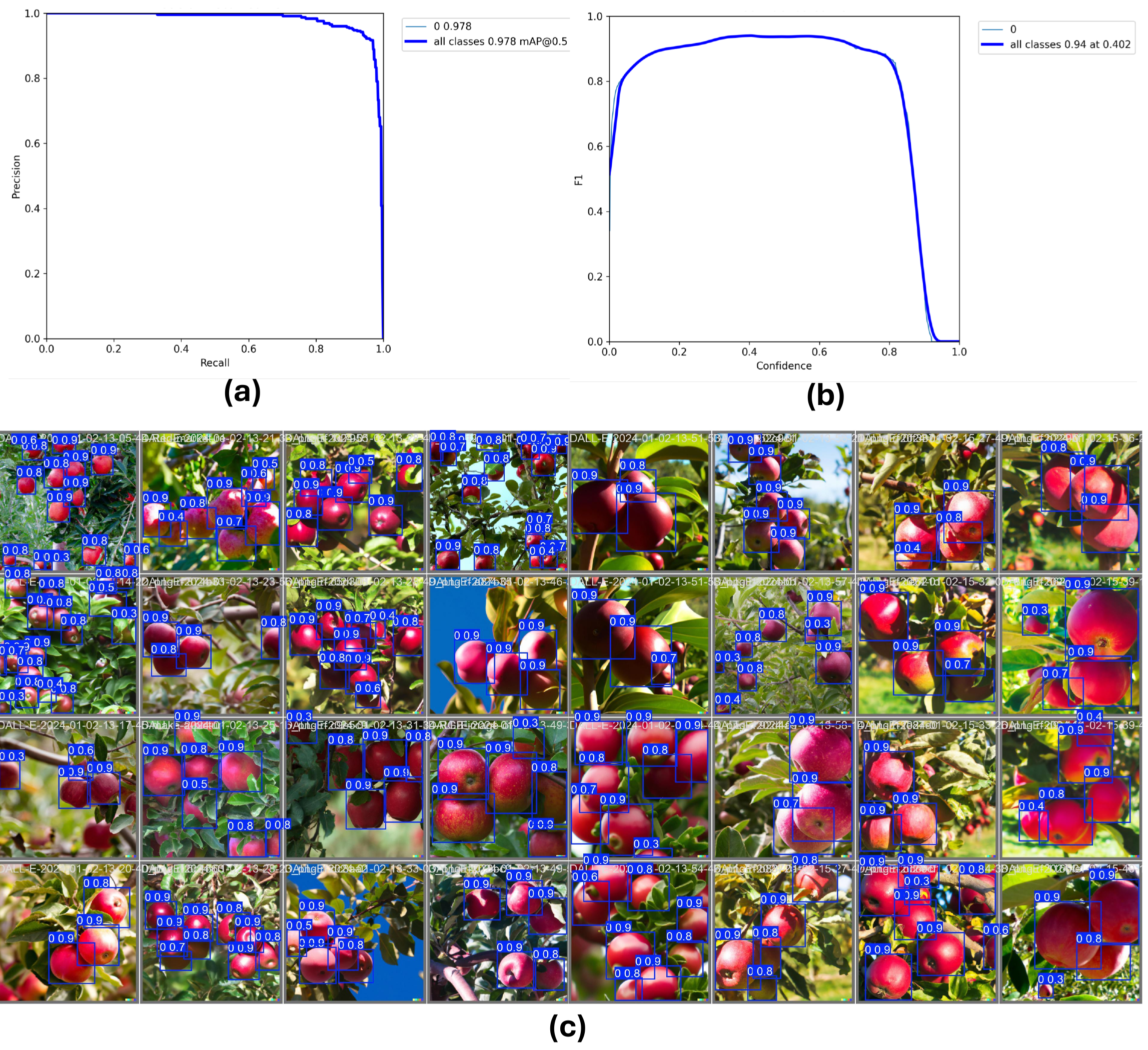}
\caption{a) Precision-Recall curve for YOLOv12n (Highest Performing Model out of YOLOv10, YOLOv11 and YOLOv12) showing superior detection accuracy. b) F1-score versus confidence level for YOLOv12n, indicating optimal threshold settings. c) Detection examples from DALL·E-generated images, highlighting YOLOv12n's effective apple recognition. The synthetic images, produced using LLMs like DALL·E \cite{sapkota2024synthetic, sapkota2024zero}, were incorporated into training datasets for YOLOv10 and YOLOv11 to mitigate real-world data limitations. These images simulate diverse challenges such as variable lighting, occlusions, clustered fruit arrangements, and complex orchard backgrounds, to enhance model adaptability. By fusing synthetic and authentic data, the models achieve improved generalization, critical for deployment in smart agriculture via machine vision sensors. This hybrid approach addresses dataset scarcity and variability, enabling precise apple detection for applications like crop health monitoring, yield prediction, and automated harvesting systems. The integration of synthetic data ensures robustness across unpredictable real-world conditions, bridging the gap between lab performance and field reliability in agricultural technology}
\label{fig:exampleimag}
\end{figure*}

Figure \ref{fig:exampleimag} illustrates the performance of the YOLOv12n model, which performed the best among fifteen YOLO configurations evaluated across YOLOv12, YOLOv11, and YOLOv10 series. Figures \ref{fig:exampleimag}a and \ref{fig:exampleimag}b present the Precision-Recall curve and the F1-confidence curve, respectively, showcasing the robustness and precision of YOLOv12n in detecting apples in synthetic images. Figure \ref{fig:exampleimag}c shows a few example canopy images generated by the DALL·E LLM with apples identified by YOLOv12n. These results demonstrate the model's effectiveness in processing and recognizing complex apple canopy images. 

\begin{figure*}[ht]
\centering
\includegraphics[width=0.8\linewidth]{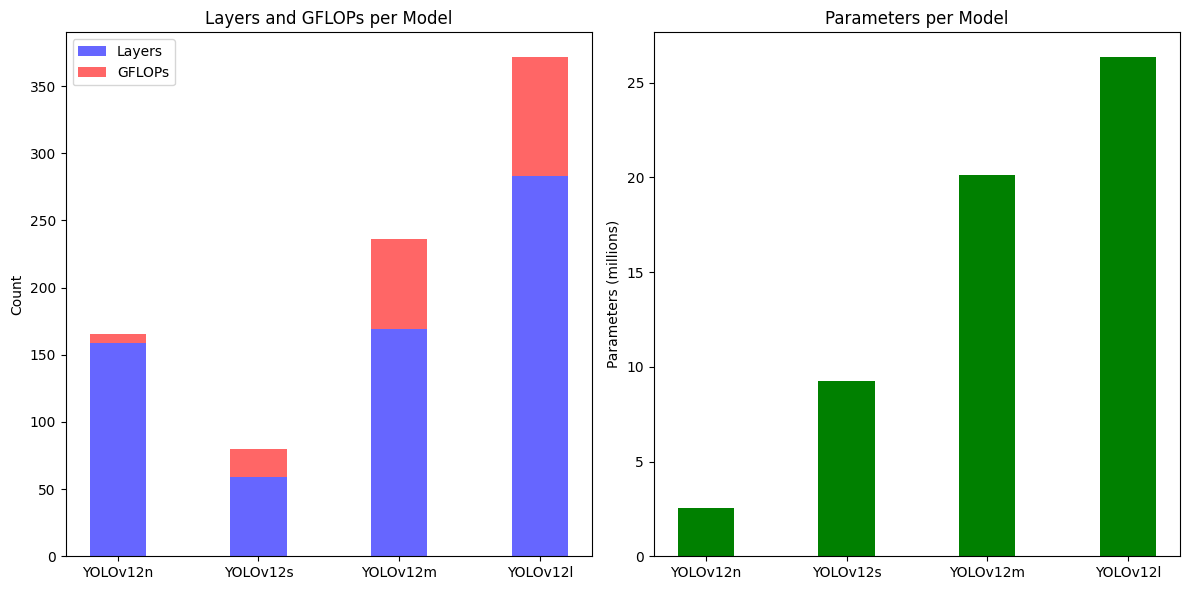}
\caption{  Left: Comparison of convolution layers and GFLOPs per YOLOv12 model. Right: Parameter count (in millions) for each model configuration.}
\label{fig:layersg}
\end{figure*}

\subsection{Evaluation of Parameters, GFLOPs and Layers used in YOLOv12}
In the comparative analysis of YOLOv12 configurations for detecting apples in synthetic orchard images, the YOLOv12n model used the fewest convolutional layers (159 layers) and demonstrated the lowest computational demand with 6.3 GFLOPs. Conversely, the YOLOv12l model, with its 283 layers and 88.5 GFLOPs, demanded the highest computational resources. Meanwhile, the YOLOv12n configuration also employed the fewest parameters at 2.556 million, indicating a leaner and more efficient architecture relative to its counterparts. Figure \ref{fig:layersg} presents the model architectural characteristics of YOLOv12n, including the number of learnable parameters and GFLOPs.

These characteristics suggest that the YOLOv12n configuration, with its minimal layer count and lower GFLOPs, offers a more practical and potentially faster deployment option for real-world applications, such as in-field apple detection using machine vision. The reduced computational load not only speeds up the inference times but also makes it more adaptable for integration into mobile or embedded systems where power and processing capabilities are limited. This efficiency can lead to broader applications and more scalable solutions in agricultural robotics and precision farming technologies.

\subsection{Evaluation of Image Processing Speed}
YOLOv12 model configurations demonstrated varying levels of computational performance. YOLOv12n model showcased the highest efficiency among all tested configurations across the YOLOv12 configurations with an inference time of only 5.6 ms. This speed is slower compared to the fastest YOLOv11 model (YOLO11n with 4.7 ms) and YOLOv10 model (YOLOv10n at 5.9 ms). As expected, increasing model complexity in the YOLOv12 series increased inference time, with YOLOv12l reaching 32.5 ms. The results showed the superior speed of YOLOv11n, compared to YOLOv12n emphasizing its potential for real-time apple detection in commercial orchards, thereby providing a scalable solution for rapid field-level image processing. 
\begin{table}[ht]
\centering
\caption{Image processing speed of YOLO model configurations across YOLOv12, YOLOv11, and YOLOv10.}
\label{tab:speeds}
\begin{tabular}{@{}p{2cm} p{1cm} p{1.5cm} p{1.5cm}@{}}
\toprule
\textbf{Model} & \textbf{Pre-process Speed (ms)} & \textbf{Inference Speed (ms)} & \textbf{Post-process Speed (ms)} \\ 
\midrule
YOLOv12n       & 0.2                             & 5.6                           & 1.4                             \\
YOLOv12s       & 0.2                             & 10.3                          & 1.0                             \\
YOLOv12m       & 0.2                             & 22.2                          & 1.2                             \\
YOLOv12l       & 0.2                             & 32.5                          & 0.5                             \\
YOLO11n        & 2.0                             & \textbf{4.7}                  & 0.6                             \\
YOLO11s        & 1.7                             & 5.6                           & 1.5                             \\
YOLO11m        & 1.9                             & 8.3                           & 0.5                             \\
YOLO11l        & 2.5                             & 9.7                           & 0.7                             \\
YOLO11x        & 2.1                             & 17.5                          & 0.5                             \\
YOLOv10n       & 1.8                             & 5.9                           & 1.8                             \\
YOLOv10s       & 1.8                             & 9.3                           & 1.8                             \\
YOLOv10m       & 1.7                             & 16.4                          & 1.6                             \\
YOLOv10b       & 1.7                             & 19.6                          & 1.6                             \\
YOLOv10l       & 1.7                             & 23.3                          & 1.6                             \\
YOLOv10x       & 1.7                             & 36.1                          & 1.6                             \\
\bottomrule
\end{tabular}
\end{table}

\subsection{Field-Testing with Real Images}
\begin{figure*}[ht]
\centering
\includegraphics[width=0.7\linewidth]{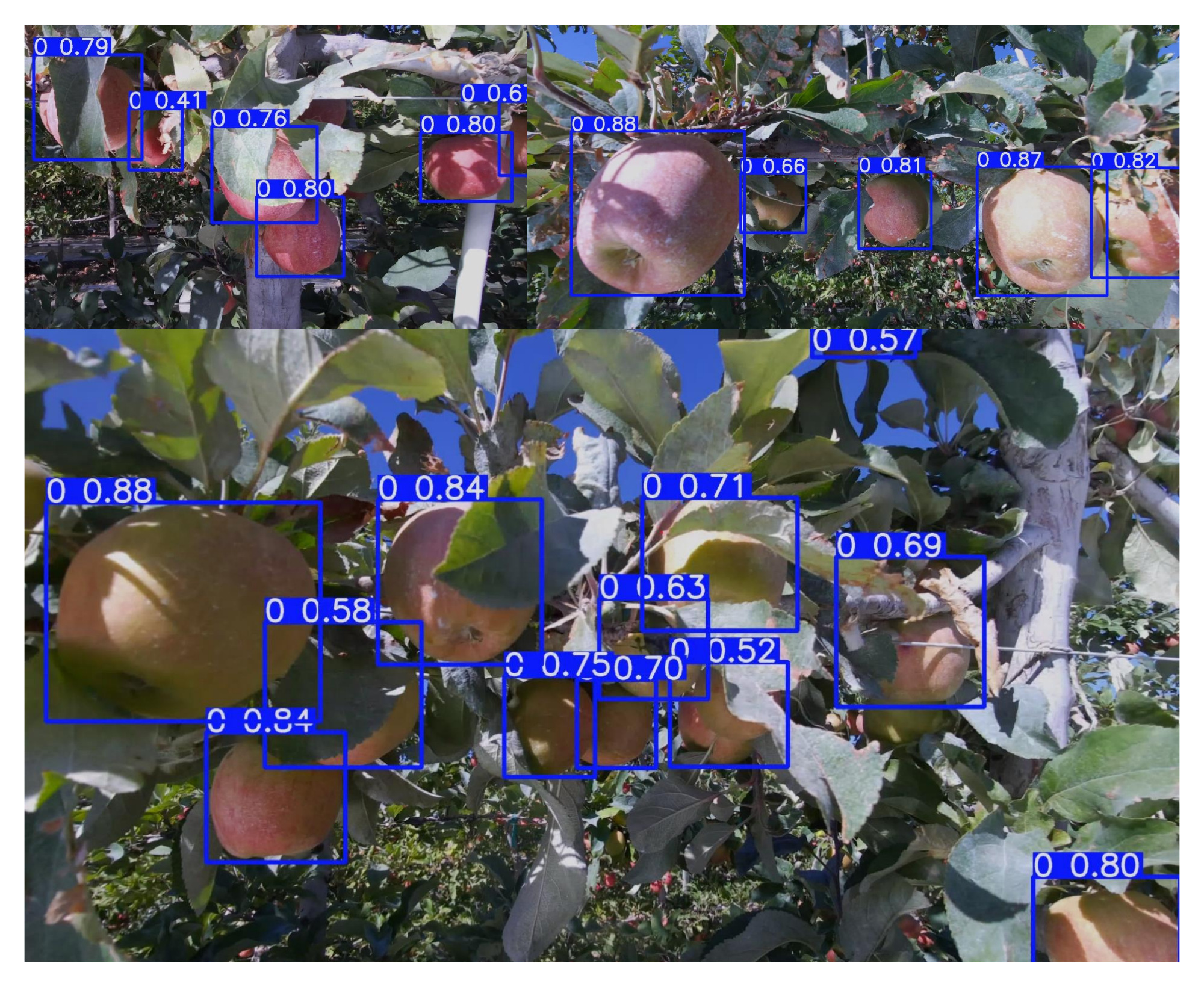}
\caption{Field image captured by Microsoft Azure Kinect camera showcasing robust apple detection using the YOLOv12n model during commercial orchard operations.}
\label{fig:fieldExample}
\end{figure*}

The robustness of the YOLOv12 model, trained exclusively on images generated by LLMs, was further validated through testing with actual field images. When deployed for real-time detection in commercial apple orchards, the model exhibited exceptional accuracy in recognizing apples from images captured by a Microsoft Azure Kinect camera mounted on a robotic ground platform, as depicted in Figure \ref{fig:fieldExample}. This validation phase, conducted during the harvest season, confirmed the model's capability to generalize from synthetic to real-world environments effectively. It was also found that the YOLOv12 model outperformed its predecessors, YOLOv11 and YOLOv10, in detecting real apples under field conditions. Our YOLOv12n model achieved an mAP@50 of 0.978, surpassing the performance metrics of all previously discussed methods. The YOLOv12s, YOLOv12m, and YOLOv12l configurations also demonstrated high precision and recall, consistently above 0.898 and 0.956 respectively. These results underscore the effectiveness of synthetic data for training. aDDITIONALLY, this findings further substantiated our earlier findings from \cite{sapkota4941582synthetic} demonstrated a significant shift in deep-learning for agricultural applications where limited to no actual field images would be necessary for model training. This approach not only reduces the time and resources typically needed for extensive data collection but also enhances the scalability of deploying AI solutions in variable agricultural environments. 

\section{Conclusion}
This study demonstrated the superior performance of the YOLOv12 model in agricultural object detection, particularly in apple orchards. The YOLOv12n configuration was found to be the best model among 15 configurations tested across YOLOv12, YOLOv11 and YOLOv10, achieving the highest performance metrics, including a box precision of 0.916, box recall of 0.969, and mAP@50 of 0.978. These results not only demonstrated the capabilities of the latest iteration of the YOLO series but also highlighted its enhancements over previous versions. Comparatively, the best-performing model in the YOLOv11 series was YOLO11x, with a precision, recall, and mAP@50 of 0.857, 0.85, and 0.91, respectively, whereas the best among the YOLOv10 series was YOLOv10n, achieving respective metrics of 0.84, 0.8, and 0.89.

The study also illustrated the ability of the YOLOv12 model to effectively leverage synthetic data model generated via LLMs for robust real-world applications. High performance metrics achieved by YOLOv12 model in field-level validation (against real images from a commercial apple orchard) indicate its practical viability and robustness in real-world applications for agricultural monitoring and automation. This advancement is for improving the scalability and robustness of the deep learning models in agricultural applications by reducing the need for extensive manual data collection and broadening the variability and size of training datasets. 

This method not only proves feasible but also highly effective in enhancing detection accuracies and inference speeds, as demonstrated by YOLOv12's superior performance in our study. As advancements continue, the fusion of more realistic LLM-generated images with the evolving YOLO series promises to expand applications beyond traditional domains, potentially transforming fields like precision agriculture

\section{Acknowledgment and Funding}
This research is funded by the National Science Foundation and the United States Department of Agriculture, National Institute of Food and Agriculture through the “AI Institute for Agriculture” Program (Award No.AWD003473). We acknowledge Achyut Paudel, Safal Kshetri, and Martin Churuvija for their support. 
\scriptsize

Our other Research on \cite{sapkota2024immature}, \cite{sapkota2024synthetic}, \cite{sapkota2024integrating}, \cite{sapkota2024zero},  \cite{sapkota2024yolov10}, \cite{meng2025yolov10}, \cite{sapkota2023creating}, \cite{sapkota2024multi}, \cite{churuvija2025pose}, \cite{sapkota2024comparing}, \cite{khanal2023machine}, \cite{sapkota2025comprehensive}, \cite{sapkota2025comprehensiveanalysistransparencyaccessibility}, \cite{sapkota2025yolov12genesisdecadalcomprehensive}.    

\bibliography{ifacconf}             
\end{document}